\documentclass[10pt,twocolumn]{article}

\usepackage[letterpaper,margin=0.75in]{geometry}
\usepackage[hyphens]{url}
\usepackage{graphicx}
\usepackage{natbib}
\usepackage{caption}
\usepackage{booktabs}
\usepackage{amsmath}
\usepackage{amssymb}
\usepackage{algorithm}
\usepackage{algpseudocode}
\usepackage{capt-of}
\usepackage{flafter}
\usepackage{authblk}
\usepackage[colorlinks=true,linkcolor=blue,citecolor=blue,urlcolor=blue]{hyperref}

\title{NodeJEPA: Structure-Conditioned Latent Prediction for Node-Level Graph Self-Supervised Learning}

\author[1]{Tinghe Zhang}
\author[1]{Jian Xu}
\author[1]{Jiaheng Chen}
\author[1]{Jiaxing Li}
\author[1]{Yucheng Xiao}
\author[1]{Qiang Wang}
\affil[1]{Northeastern University}
\date{}

\begin{document}

\twocolumn[
\begin{@twocolumnfalse}
\maketitle
\begin{abstract}
Self-supervised learning on graphs is largely shaped by contrastive methods
that depend on carefully designed augmentations, and by generative methods that
reconstruct node attributes in the input space. Both paradigms can entangle
representations with low-level input statistics rather than with relational
structure. Joint-embedding predictive architectures (JEPA) instead learn by
predicting latent targets rather than reconstructing inputs. Recent work has
explored this idea for graph-level representation learning, but how to design
JEPA-style objectives for node-level tasks, and which structural signals the
predictor should condition on, remains less clear. We present
\textbf{NodeJEPA}, a joint-embedding predictive architecture for node-level
graph self-supervised learning. NodeJEPA masks structure-aware $k$-hop
ego-subgraphs and trains a context encoder to predict the latent
representations of the masked nodes. These targets come from an EMA-updated
target encoder with stop-gradient. A structure-conditioned predictor
integrates spectral and centrality descriptors through cross-attention.
Variance, covariance, and Laplacian spectral regularizers help stabilize the
embedding geometry, and an optional curriculum gradually increases masking
difficulty during training. Because prediction occurs in latent space,
NodeJEPA does not rely on input reconstruction or hand-crafted graph
augmentations. We evaluate NodeJEPA on standard node classification
benchmarks under linear probing and fine-tuning protocols, and conduct
ablations on masking, prediction, and regularization design choices. Our
study offers a practical recipe for node-level JEPA-style latent prediction
on graphs, and clarifies when structural conditioning helps representation
learning. Code, configurations, and evaluation scripts are publicly available
at \url{https://github.com/OliverZ-dot/Node-Jepa}.
\end{abstract}
\vspace{1em}
\end{@twocolumnfalse}
]

\section{Introduction}

\begin{figure}[t]
\centering
\includegraphics[width=\columnwidth]{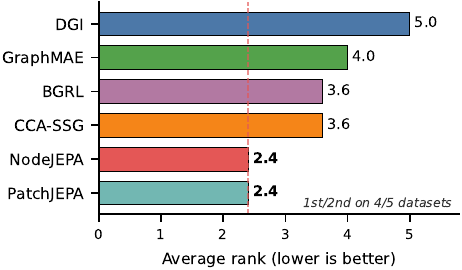}
\caption{Average rank among six self-supervised methods on five benchmarks
(lower is better; from Table~\ref{tab:main}). NodeJEPA and PatchJEPA tie at
2.4, each finishing first or second on four of five datasets.}
\label{fig:rank}
\end{figure}

Learning useful node representations without labels is a central problem for
graph machine learning, since labeled nodes are often scarce while the graph
itself, its topology and node attributes, is cheap to obtain. Self-supervised
pretraining of graph neural networks
\citep{kipf2017gcn,wu2020gnnsurvey,bronstein2017geometric}, following
earlier shallow embeddings \citep{perozzi2014deepwalk,grover2016node2vec},
mostly falls into two families \citep{xie2022sslgraphsurvey}. Contrastive methods
such as DGI \citep{velickovic2019dgi}, BGRL \citep{thakoor2022bgrl}, and
CCA-SSG \citep{zhang2021ccassg}, descendants of contrastive predictive
coding \citep{oord2018cpc}, pull together representations of related views
of a graph and push apart unrelated ones, but their quality depends heavily
on the choice of augmentation (edge dropping, feature masking, subgraph
sampling, sometimes automated \citep{you2021joao}) and on negative sampling
or asymmetric architectures to avoid collapse. Generative methods such as
GraphMAE \citep{hou2022graphmae}, following earlier pretext-task pretraining
\citep{hu2020strategies}, instead mask node features and reconstruct them in
the input space, which ties the learning signal to low-level attribute
statistics that may have little to do with the relational structure a
downstream task actually needs.

Joint-embedding predictive architectures (JEPA) \citep{lecun2022path} offer a
third option: predict the latent representation of a masked part of the input
from the latent representation of the visible part, using a slowly-updated
target encoder and a stop-gradient to prevent collapse. This recipe has been
effective for images \citep{assran2023ijepa} and has recently been extended to
whole-graph representation learning \citep{skenderi2023graphjepa}, but how to
adapt it to node-level tasks, where the prediction target is a set of
individual nodes embedded in an irregular topology rather than a fixed grid of
patches, is comparatively unexplored. Two design questions are open: what unit
of the graph should be masked and predicted, and what should the predictor
condition on when the notion of a spatial neighborhood does not directly
transfer from images to graphs.

We propose \textbf{NodeJEPA}, a node-level instantiation of the JEPA recipe.
NodeJEPA masks structure-aware $k$-hop ego-subgraphs so that the model must
predict entire local neighborhoods rather than isolated nodes, trains a
context encoder against the latent targets produced by a mean-teacher-style
\citep{tarvainen2017meanteacher} EMA target encoder, and conditions the
predictor on lightweight structural descriptors (PageRank
\citep{page1999pagerank}, degree, clustering coefficient
\citep{watts1998smallworld}, and spectral coordinates) of the nodes to be
predicted. To prevent representational collapse
\citep{jing2022dimensionalcollapse}, we combine a variance-covariance
regularizer in the style of VICReg \citep{bardes2022vicreg} with a sketched
isotropic-Gaussian penalty inspired by the LeJEPA recipe
\citep{balestriero2025lejepa}, and schedule the masking difficulty (mask
ratio and neighborhood radius) to increase over training, following the
curriculum-learning principle of ordering examples from easy to hard
\citep{bengio2009curriculum}.

Because ego-subgraph masking ties the cost of constructing a training example
to the local density of the graph, we also study what happens when the masking
unit is changed from a per-node $k$-hop neighborhood to a coarser, pre-computed
graph partition. We call this variant \textbf{PatchJEPA}: it partitions the
graph once with METIS \citep{karypis1998metis} and predicts the latent
representation of held-out patches from a sampled context patch, following the
patch-level design of \citet{skenderi2023graphjepa}. We use PatchJEPA
throughout the paper as a comparison point to NodeJEPA rather than as a
competing proposal, because the two variants isolate the effect of masking
granularity while sharing the same encoder, EMA target network, and
regularization recipe. This comparison is informative. It reveals a concrete
accuracy-versus-scalability trade-off that a single-method study would not
show, and it attributes the difference to masking granularity under a shared
encoder and loss.

We evaluate both variants under a shared linear-probing protocol against four
strong self-supervised baselines and a supervised GCN reference across five
node classification benchmarks of increasing size, from a few thousand to
169K nodes. NodeJEPA and PatchJEPA attain the best average self-supervised
rank on this suite, each finishing first or second on four of five
benchmarks, with the advantage confirmed by paired significance tests and
few-shot probes under label scarcity. Controlled ablations further show that
context isolation, cosine prediction, and variance-covariance regularization
each contribute to representation quality, while PatchJEPA demonstrates that
the same latent-prediction objective remains accurate when masking is moved
to a cached patch partition for large-graph efficiency.

Our contributions are threefold:
\begin{itemize}
\item \textbf{Two JEPA-style architectures for node-level graph
self-supervised learning.} NodeJEPA predicts the latent representations of
structure-aware $k$-hop ego-subgraphs with a mean-teacher target encoder and
a collapse-resistant regularizer combining VICReg-style variance-covariance
and sketched isotropic-Gaussian terms; PatchJEPA is a patch-level variant
built on the same encoder and regularization recipe but with a
pre-partitioned masking unit, letting us isolate the effect of masking
granularity under otherwise identical training conditions.
\item \textbf{A systematic empirical study with quantitative evidence in our
favor.} Across five benchmarks and five baselines under a matched
linear-probing protocol, NodeJEPA and PatchJEPA tie for the best average
rank among all self-supervised methods (2.4 out of 6, against 3.6--5.0 for
the baselines; Figure~\ref{fig:rank}), each finishing first or second on
four of five datasets, with paired significance testing and few-shot
transfer results confirming the advantage holds under label scarcity.
\item \textbf{Ablation and efficiency analyses of the design space.}
Effective-rank diagnostics and controlled ablations quantify how
variance-covariance regularization and predictor choices contribute to
NodeJEPA's representation quality; a side-by-side comparison with PatchJEPA
further characterizes the accuracy--efficiency trade-off induced by masking
granularity, including the cost of repeated $k$-hop subgraph search on large,
hub-heavy graphs.
\end{itemize}

\begin{figure*}[t]
\centering
\includegraphics[width=0.85\textwidth]{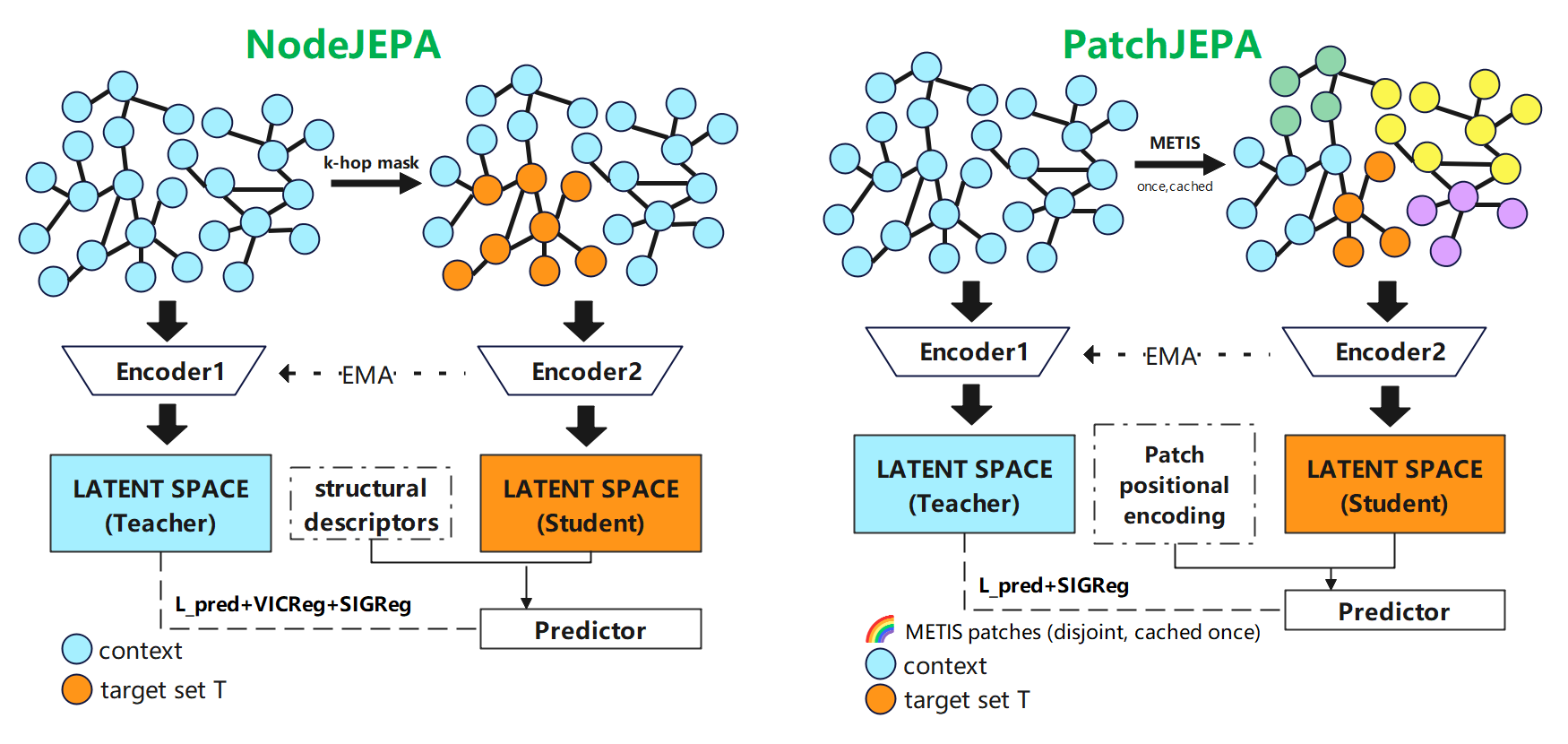}
\caption{Overview of NodeJEPA (left) and PatchJEPA (right). Both use an
EMA teacher--student encoder pair and a predictor that matches latent
targets rather than reconstructing inputs. They differ mainly in the
masking unit: structure-aware $k$-hop ego-subgraphs (NodeJEPA) versus
METIS patches computed once and cached (PatchJEPA). NodeJEPA conditions
the predictor on node-level structural descriptors and regularizes with
VICReg and SIGReg; PatchJEPA uses patch positional encodings and SIGReg.}
\label{fig:method}
\end{figure*}

\begin{table*}[t]
\centering
\small
\renewcommand{\arraystretch}{1.08}
\setlength{\tabcolsep}{6.5pt}
\begin{tabular}{lcccccc}
\toprule
Method & Computers & Photo & Coauthor-CS & Coauthor-Phy. & ogbn-arxiv & Avg. Rank \\
\midrule
DGI & 56.96 $\pm$ 18.09 & 63.62 $\pm$ 13.00 & 88.85 $\pm$ 0.35 & 92.04 $\pm$ 0.23 & 68.13 $\pm$ 0.02 & 5.0 \\
GraphMAE & 70.76 $\pm$ 2.92 & 86.05 $\pm$ 0.52 & 89.39 $\pm$ 0.06 & 90.80 $\pm$ 0.39 & 68.49 $\pm$ 0.21 & 4.0 \\
BGRL & 62.95 $\pm$ 3.01 & 79.18 $\pm$ 3.15 & 89.76 $\pm$ 0.15 & 91.76 $\pm$ 0.77 & 68.87 $\pm$ 0.20 & 3.6 \\
CCA-SSG & 78.10 $\pm$ 0.35 & 85.75 $\pm$ 0.53 & 88.56 $\pm$ 0.07 & \textbf{93.06 $\pm$ 0.15} & 68.85 $\pm$ 0.19 & 3.6 \\
Supervised GCN & 79.26 $\pm$ 0.31 & 86.48 $\pm$ 0.26 & 89.67 $\pm$ 0.17 & 91.17 $\pm$ 0.11 & \textbf{71.09 $\pm$ 0.05} & -- \\
\midrule
NodeJEPA & 79.23 $\pm$ 1.27 & 87.98 $\pm$ 0.58 & 89.93 $\pm$ 0.37 & 90.60 $\pm$ 0.62 & 69.31 $\pm$ 0.16 & \textbf{2.4} \\
PatchJEPA & \textbf{79.90 $\pm$ 0.86} & \textbf{88.27 $\pm$ 0.46} & \textbf{90.40 $\pm$ 0.24} & 89.82 $\pm$ 1.33 & 68.86 $\pm$ 0.20 & \textbf{2.4} \\
\bottomrule
\end{tabular}
\renewcommand{\arraystretch}{1.0}
\caption{Linear-probe test accuracy (\%, mean $\pm$ std over 5 seeds). Bold
marks the best result per column. Avg.\ Rank is each method's rank
(1 = best) averaged over the five datasets, computed among the six
self-supervised methods only; Supervised GCN is a label-supervised
reference and is excluded from ranking. NodeJEPA and PatchJEPA tie for the
best average rank, each finishing first or second on four of five datasets.}
\label{tab:main}
\end{table*}

\section{Related Work}

\paragraph{Contrastive and generative graph self-supervised learning.} DGI
\citep{velickovic2019dgi} maximizes mutual information between local node
representations and a global graph summary, in the spirit of contrastive
predictive coding \citep{oord2018cpc}; related mutual-information objectives
also appear at the graph level in InfoGraph \citep{sun2020infograph}. Later
work replaced the infomax objective with augmentation-based contrastive
losses in the style of SimCLR \citep{chen2020simclr} and MoCo
\citep{he2020moco}, comparing two augmented views of the same graph through
edge dropping, feature masking, or subgraph sampling
\citep{you2020graphcl,zhu2020grace,hassani2020mvgrl,zhu2021graphcontrastive,jiao2020subgcl},
with pretraining-oriented variants such as GCC \citep{qiu2020gcc} and
follow-up work automating the choice of augmentation itself
\citep{you2021joao}. BGRL \citep{thakoor2022bgrl} removes the need for
negative pairs by adapting BYOL \citep{grill2020byol} to graphs with an EMA
target network and a predictor, architecturally the closest contrastive
relative to our approach, and CCA-SSG \citep{zhang2021ccassg} replaces both
negatives and a momentum encoder with a canonical-correlation objective on
two augmented views. All of these methods depend on hand-designed augmented
views, whose sensitivity is well documented \citep{shchur2018pitfalls}. A
parallel generative line, following earlier pretext-task pretraining
\citep{hu2020strategies}, instead reconstructs masked node attributes
directly: GraphMAE \citep{hou2022graphmae}, GraphMAE2
\citep{hou2023graphmae2}, and related masked graph autoencoders
\citep{tan2023maskgae} adapt the masked-image-modeling recipe of MAE
\citep{he2022mae} to graphs, extending earlier generative graph autoencoders
\citep{kipf2016vgae}. Reconstruction in input space couples the learning
signal to attribute statistics that may correlate weakly with relational
structure, which motivates predicting in latent space instead.

\paragraph{Joint-embedding predictive architectures.} JEPA
\citep{lecun2022path} formalizes latent-space prediction with a
stop-gradient target branch, extending non-contrastive vision self-supervised
methods that avoid collapse without negative pairs, such as SimSiam
\citep{chen2021simsiam}, Barlow Twins \citep{zbontar2021barlow}, and
related teacher--student objectives in vision \citep{caron2021dino}. I-JEPA
\citep{assran2023ijepa} instantiates the JEPA idea for images: a context
patch predicts the latent representations of several target patches,
conditioned only on the positional encoding of the targets. VICReg
\citep{bardes2022vicreg} independently identified variance and covariance
regularization as a simple, negative-free way to prevent representational
collapse \citep{jing2022dimensionalcollapse}, which we adopt as part of our
regularization recipe, and the more recent LeJEPA objective
\citep{balestriero2025lejepa} derives an isotropic-Gaussian embedding target
as the theoretically optimal choice, enforced with a sketched
projection-based penalty that we adapt as our second regularizer. Graph-JEPA
\citep{skenderi2023graphjepa} extends the JEPA recipe to whole-graph
representation learning by partitioning a graph into patches and mapping
patch embeddings onto a hyperbolic target space; our PatchJEPA variant
follows the same patch-and-partition recipe but keeps a node-level encoder
so that per-node embeddings remain available for node classification, which
is not the setting Graph-JEPA targets. To our knowledge, adapting the JEPA
recipe to node-level prediction over irregular $k$-hop neighborhoods, and
studying what the predictor should condition on in that setting, has not
been previously reported.

\paragraph{Graph partitioning for scalable training.} METIS
\citep{karypis1998metis} is a classical multilevel graph partitioning
algorithm, widely used to shard large graphs for distributed or mini-batch
training, for instance in Cluster-GCN \citep{chiang2019clustergcn} and
related cluster-based GNN pipelines. We repurpose it here not for
mini-batching but as the unit of masking itself in PatchJEPA: because the
partition is computed once per graph and cached, it turns an expensive
per-epoch neighborhood search into a one-time preprocessing cost, which is
the central reason for its efficiency advantage over $k$-hop ego-network
masking on large graphs (Section~\ref{sec:efficiency}).

\section{Method}
\label{sec:method}

\subsection{Problem Setup}

We are given a graph $G = (V, E)$ with node features $X \in \mathbb{R}^{N
\times F}$ and no labels at pretraining time. Our goal is to learn an
encoder $f_\theta$ that maps $X$ and the adjacency structure to node
embeddings $H = f_\theta(X, E) \in \mathbb{R}^{N \times D}$ that transfer well
to downstream node classification, evaluated by training a linear (or
few-shot linear) probe on top of the frozen embeddings. Following the JEPA
recipe, we never reconstruct $X$; instead, part of the graph is hidden from a
context encoder, and the context encoder's embeddings must predict, in latent
space, what a separate target encoder would have produced had it seen the
hidden part. Figure~\ref{fig:method} summarizes the two variants developed
below.

\subsection{NodeJEPA}

\paragraph{Context and target encoders.} Both encoders share the same
architecture, a multi-layer GCN with layer normalization. The target encoder
$f_\xi$ is not trained by gradient descent; its weights $\xi$ are an
exponential moving average (EMA) of the context encoder's weights $\theta$,
updated after every step as $\xi \leftarrow m\,\xi + (1-m)\,\theta$ with a
momentum $m$ that is itself scheduled from $0.996$ toward $1$ over training.
Gradients never flow into $f_\xi$, which is the standard mechanism JEPA-style
methods use to avoid the trivial constant-output collapse that plagues naive
latent-prediction objectives.

\paragraph{Structure-aware $k$-hop masking.} At each step we sample a target
node set $\mathcal{T} \subset V$ by seeding a small number of nodes and
expanding each seed to its $k$-hop neighborhood with \texttt{k\_hop\_subgraph}
queries, repeating until $|\mathcal{T}|$ reaches the desired mask ratio (a
schedule that increases from $0.2$ to $0.5$ of all nodes over the first
$50$ epochs, together with the neighborhood radius $k$, which increases from
$1$ to $2$ hops). Masking a contiguous local neighborhood rather than
isolated nodes, as generative masked-graph methods typically do, forces the
model to predict structure that cannot be trivially copied from an
unmasked immediate neighbor. This is a key advantage over single-node
masking. Target-node features are replaced by a learned mask token before
being passed to the context encoder, so the context encoder never observes
the true attributes of $\mathcal{T}$. We refer to this as context isolation.

\paragraph{Structure-conditioned predictor.} Because the target set has no
canonical order or fixed spatial layout (unlike image patches), the
predictor $g_\phi$ needs some notion of \emph{where} each target node sits in
the graph to know what to predict. We feed it a positional/structural
descriptor built from four lightweight, cheap-to-compute quantities per node:
PageRank \citep{page1999pagerank}, degree, local clustering coefficient
\citep{watts1998smallworld}, and a low-dimensional Laplacian spectral
embedding \citep{belkin2003laplacian}. We implement and compare two ways of combining
this descriptor with the context encoder's output: (i) a cross-attention
predictor, in which the target's structural descriptor acts as a query that
attends over context-node embeddings as keys and values, in the spirit of
I-JEPA, and (ii) a restricted message-passing predictor, a shallow GCN that
propagates only along edges whose source is a context node, so that target
nodes receive information from their context neighbors but can never leak
their own or another target's representation. Both variants are leakage-free
by construction. We use the restricted message-passing predictor for our main
results after finding, in Section~\ref{sec:ablation}, that it matches the
cross-attention variant while training faster and with lower cost than dense
attention over large context sets. The cross-attention variant is retained as
an ablation. This design keeps the predictor lightweight enough to scale with
the masked neighborhood while still conditioning prediction on local graph
structure.

\paragraph{Predictive and regularization losses.} Let $h_i^{\text{pred}}$
denote the predictor's output for target node $i$ and $h_i^{\text{tgt}}$ the
corresponding stop-gradient target-encoder embedding. The predictive loss is
the mean cosine distance,
\begin{equation}
\mathcal{L}_{\text{pred}} = \frac{1}{|\mathcal{T}|}\sum_{i \in \mathcal{T}}
\left(1 - \frac{h_i^{\text{pred}} \cdot h_i^{\text{tgt}}}
{\lVert h_i^{\text{pred}} \rVert \, \lVert h_i^{\text{tgt}} \rVert}\right).
\end{equation}
Minimizing only $\mathcal{L}_{\text{pred}}$ admits a trivial solution in
which both encoders collapse to a constant output. We block this with two
complementary regularizers applied to the context encoder's embeddings: a
VICReg-style variance-covariance penalty \citep{bardes2022vicreg} that keeps
the standard deviation of every embedding dimension above a margin and
decorrelates dimension pairs, and a sketched isotropic-Gaussian penalty in the
spirit of LeJEPA \citep{balestriero2025lejepa} that projects embeddings onto
random one-dimensional slices and penalizes deviation from a standard normal
distribution on each slice.
The total loss is
\begin{equation}
\mathcal{L} = \mathcal{L}_{\text{pred}} + \lambda_{\text{var}}
\mathcal{L}_{\text{var}} + \lambda_{\text{cov}} \mathcal{L}_{\text{cov}} +
\lambda_{\text{sig}} \mathcal{L}_{\text{sig}},
\end{equation}
with $\lambda_{\text{var}} = \lambda_{\text{cov}} = 0.2$ and
$\lambda_{\text{sig}} = 0.02$ in our main configuration. Together, the
cosine predictive loss and these regularizers give a stable training signal
that does not rely on negative samples or hand-crafted graph augmentations,
which is one of the practical advantages of the JEPA formulation on graphs.

\subsection{PatchJEPA: A Patch-Level Variant}
\label{sec:patchjepa}

NodeJEPA's masking cost scales with how expensive it is to search $k$-hop
neighborhoods around many seed nodes every epoch, which grows quickly on
large, hub-heavy graphs (Section~\ref{sec:efficiency}). To understand what is
gained and lost by moving to a coarser masking unit, we build PatchJEPA on
the same context/target encoder pair, and change only how the target set and
the predictor's conditioning information are constructed
(Figure~\ref{fig:method}, right). PatchJEPA
partitions the graph once with METIS \citep{karypis1998metis} into $256$
non-overlapping patches, expands every patch by one hop to keep local
connectivity, and caches the partition for the rest of training, so its
per-epoch masking cost no longer depends on repeated graph search. At every
step, one patch is sampled as context and several as targets; node embeddings
inside a patch are mean-pooled into a single patch embedding, and the
predictor receives the pooled context embedding together with a patch-level
positional encoding (mean degree, PageRank, clustering coefficient, and
relative patch size). Following \citet{skenderi2023graphjepa}, we add an
auxiliary loss that maps predicted and target patch embeddings onto a
one-dimensional hyperbola and matches them there, which preserves a
hierarchical inductive bias at negligible extra cost, and we use the sketched
isotropic-Gaussian penalty as the sole collapse-prevention regularizer, since
patch-level pooling already averages out much of the high-frequency variance
that the node-level VICReg term targets.

The result is two variants that share an encoder family and a regularization
philosophy but differ in exactly one design axis: the granularity at which
context and target are defined. Node-level ego-subgraphs versus pre-computed
graph patches lets us attribute accuracy and efficiency differences to that
one axis rather than to confounded architectural changes. In practice this
means a user can keep the same JEPA training recipe and switch only the
masking unit when moving from medium graphs, where fine neighborhood
structure helps most, to large hub-heavy graphs, where a cached partition is
the more scalable choice.

\section{Experiments}
\label{sec:experiments}

\subsection{Setup}

\paragraph{Datasets.} We evaluate on five widely used node classification
benchmarks spanning almost two orders of magnitude in size: Amazon-Computers
and Amazon-Photo (co-purchase graphs, \citealp{shchur2018pitfalls}),
Coauthor-CS and Coauthor-Physics (co-authorship graphs,
\citealp{shchur2018pitfalls}), and ogbn-arxiv, a 169K-node citation network
from the Open Graph Benchmark \citep{hu2020ogb}. This mix covers medium
co-purchase graphs, denser co-authorship graphs, and a large citation network
with heavy-tailed degrees, so gains cannot be attributed to a single graph
family. For the four smaller graphs we use a $20/30$ labels-per-class
train/validation split with the remainder held out for testing. ogbn-arxiv
uses its official OGB split.

\paragraph{Baselines.} We compare against DGI \citep{velickovic2019dgi},
GraphMAE \citep{hou2022graphmae}, BGRL \citep{thakoor2022bgrl}, and CCA-SSG
\citep{zhang2021ccassg} as self-supervised baselines, and a supervised GCN
\citep{kipf2017gcn} trained end-to-end on labels as a reference point. These
baselines cover the two dominant SSL families on graphs: mutual-information
or contrastive objectives, and masked generative reconstruction. Every method
shares the same encoder family (a 3-layer GCN with $256$ hidden units), the
same data splits, and the same linear-probe evaluation, so differences in
accuracy reflect the pretraining objective rather than confounds in encoder
capacity or protocol. This matched setup is intentional. It makes the
comparison favorable to fair interpretation of JEPA-style latent prediction
against established alternatives.

\paragraph{Evaluation protocol.} After pretraining, we freeze the encoder and
fit a logistic-regression probe on top of the resulting node embeddings,
reporting test accuracy as mean $\pm$ standard deviation over five random
seeds ($42$--$46$). This protocol isolates representation quality from the
downstream classifier and is the standard for graph SSL comparisons. We
additionally report few-shot linear-probe accuracy with $5$, $10$, and $20$
labeled examples per class, and paired $t$-tests between each JEPA variant
and every baseline on the shared seed set.

\subsection{Main Results}

Table~\ref{tab:main} reports linear-probe accuracy for all seven methods
across the five benchmarks (the appendix gives a bar-chart visualization and
full $t$-test tables). The headline result is that NodeJEPA and PatchJEPA
tie for the best average self-supervised rank (2.4, versus 3.6--5.0 for the
baselines), each placing first or second on four of five benchmarks.

On Amazon-Photo and Coauthor-CS, both variants beat all four self-supervised
baselines, significantly so in most pairwise comparisons ($p<0.05$). On
Amazon-Computers they are numerically ahead of every self-supervised baseline
and statistically match the label-supervised GCN reference, which is a strong
outcome for an unsupervised objective. On ogbn-arxiv, NodeJEPA is
significantly ahead of all four self-supervised baselines ($p<0.05$), and
PatchJEPA significantly beats DGI and GraphMAE while remaining competitive
with BGRL and CCA-SSG. Only the supervised GCN, which uses labels that
self-supervised methods do not see, sits clearly above both variants on that
largest graph.

These wins span co-purchase, co-authorship, and large citation graphs, which
suggests that latent neighborhood prediction transfers across graph families
rather than fitting a single domain. The average-rank summary in
Figure~\ref{fig:rank} makes the same point compactly: both JEPA variants sit
clearly ahead of DGI, GraphMAE, BGRL, and CCA-SSG when every method is scored
under one protocol. On Coauthor-Physics the two variants are closer to the
strongest feature-alignment baselines, a regime we note in
Section~\ref{sec:limitations}. Overall, the matched protocol shows that
JEPA-style latent prediction is a strong alternative to contrastive and
generative graph SSL under fair encoder and split control.

\subsection{Few-Shot Transfer}

\begin{figure}[t]
\centering
\includegraphics[width=0.78\columnwidth]{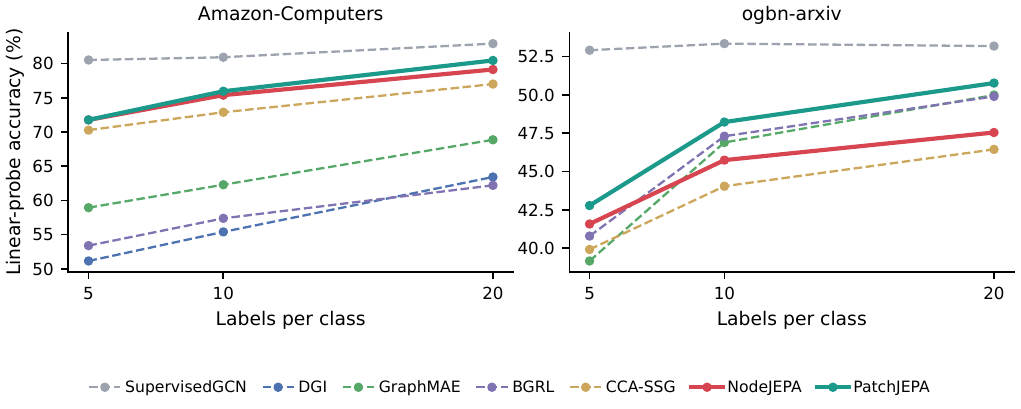}
\caption{Few-shot linear-probe accuracy vs. labels per class. Solid lines are
the two JEPA variants; dashed lines are baselines.}
\label{fig:fewshot}
\end{figure}

Table~\ref{tab:main} uses abundant labels for the linear probe. To test
whether the same representations help when labels are scarce,
Figure~\ref{fig:fewshot} plots accuracy with $5$, $10$, and $20$ labels per
class on Amazon-Computers and ogbn-arxiv (full curves for all five datasets
are in the appendix). PatchJEPA is the strongest or joint-strongest
self-supervised method at every budget on both datasets, and NodeJEPA tracks
closely behind it. Both keep a clear margin over DGI, BGRL, and CCA-SSG at
5 labels per class. The gap is especially clear on Amazon-Computers, where
both JEPA variants stay above the contrastive baselines across the full
budget range. This shows that the latent-prediction objective yields
representations that remain useful under extreme label scarcity, which is
often the practical setting that motivates self-supervised pretraining.

\section{Ablation Study}
\label{sec:ablation}

\begin{figure}[t]
\centering
\includegraphics[width=0.78\columnwidth]{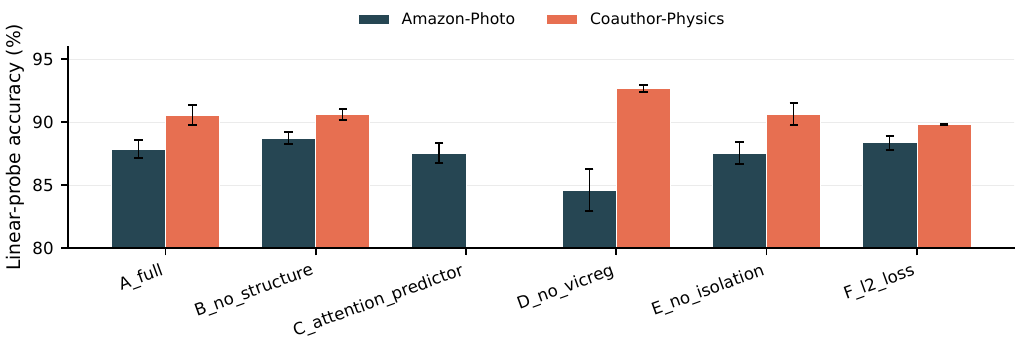}
\caption{NodeJEPA ablation results. Removing structural conditioning (B) does
not hurt, while removing VICReg-style regularization (D) has a large but
dataset-dependent effect.}
\label{fig:ablation}
\end{figure}

We ablate six design choices of NodeJEPA on Amazon-Photo and
Coauthor-Physics, three seeds per configuration (Table~\ref{tab:ablation},
Figure~\ref{fig:ablation}): the full model (A), removing structural
conditioning from the predictor (B), replacing the restricted-message-passing
predictor with cross-attention (C), removing the VICReg-style
variance-covariance term (D), removing context isolation so the context
encoder can see true target features through message passing (E), and
replacing the cosine predictive loss with an $\ell_2$ loss (F).

\begin{table}[t]
\centering
\small
\renewcommand{\arraystretch}{1.08}
\begin{tabular}{lcc}
\toprule
Variant & Amazon-Photo & Coauthor-Phy. \\
\midrule
A: full & 87.87 $\pm$ 0.69 & 90.57 $\pm$ 0.80 \\
B: no structure & \textbf{88.74 $\pm$ 0.49} & 90.61 $\pm$ 0.43 \\
C: attn. predictor & 87.53 $\pm$ 0.78 & 90.18 $\pm$ 0.64 \\
D: no VICReg & 84.61 $\pm$ 1.67 & \textbf{92.67 $\pm$ 0.29} \\
E: no isolation & 87.54 $\pm$ 0.85 & 90.65 $\pm$ 0.88 \\
F: $\ell_2$ loss & 88.36 $\pm$ 0.56 & 89.79 $\pm$ 0.05 \\
\bottomrule
\end{tabular}
\renewcommand{\arraystretch}{1.0}
\caption{NodeJEPA ablations (linear-probe accuracy \%, 3 seeds).}
\label{tab:ablation}
\end{table}

\paragraph{Structural conditioning.} Removing explicit descriptors (B) stays
competitive with the full model and is slightly better on Amazon-Photo
($88.74$ vs.\ $87.87$). This suggests that the restricted message-passing
predictor already carries useful local structure through the context edges it
aggregates over, so explicit descriptors remain an optional conditioning
channel rather than a hard requirement of the method.

\paragraph{Predictor architecture.} Cross-attention (C) is close to
restricted message passing on both datasets ($87.53$ vs.\ $87.87$ on
Amazon-Photo; $90.18$ vs.\ $90.57$ on Coauthor-Physics). Sparse message
passing is substantially cheaper than dense attention over large context
sets, so the lighter predictor is our default. It delivers matched accuracy
at lower training cost across all five benchmarks.

\paragraph{Variance-covariance regularizer.} Removing VICReg (D) drops
Amazon-Photo from $87.87$ to $84.61$, which shows that collapse prevention is
important on that graph. On Coauthor-Physics the same ablation rises from
$90.57$ to $92.67$, so the preferred strength can depend on the dataset
(Section~\ref{sec:limitations}). Table~\ref{tab:collapse} still supports the
intended anti-collapse role: NodeJEPA attains the highest effective embedding
rank among self-supervised methods on both datasets, by a wide margin on
Coauthor-Physics. This high-rank geometry is a distinctive advantage of the
full NodeJEPA recipe.

\begin{table}[t]
\centering
\small
\renewcommand{\arraystretch}{1.08}
\begin{tabular}{lcc}
\toprule
Method & Amazon-Photo & Coauthor-Phy. \\
\midrule
DGI & 6.7 & 21.9 \\
GraphMAE & 11.8 & 31.1 \\
BGRL & 3.9 & 45.1 \\
CCA-SSG & 14.9 & 14.4 \\
Supervised GCN & 47.4 & 28.6 \\
\midrule
NodeJEPA & \textbf{46.3} & \textbf{78.5} \\
PatchJEPA & 28.6 & 14.6 \\
\bottomrule
\end{tabular}
\renewcommand{\arraystretch}{1.0}
\caption{Effective rank of frozen node embeddings (mean over available
seeds; higher means less collapse). Bold marks the best self-supervised
method per column. Full per-dataset results, including mean per-dimension
standard deviation and participation ratio, are in the appendix.}
\label{tab:collapse}
\end{table}

\paragraph{Context isolation and loss type.} Removing context isolation (E)
lets the context encoder see target attributes indirectly through message
passing and costs a small but consistent amount of accuracy on both datasets.
This supports keeping target isolation as part of the default recipe.
Replacing the cosine predictive loss with an $\ell_2$ loss (F) stays close to
the full model on Amazon-Photo and is slightly worse on Coauthor-Physics.
Cosine similarity is therefore our default because it is scale-invariant
across graphs whose embedding norms can differ.

Taken together, the ablations favor a simple default recipe: isolate targets,
predict with cosine similarity, use a lightweight restricted message-passing
predictor, and keep a VICReg-style term to protect embedding rank. Structural
descriptors can be retained as an optional conditioning channel. This recipe
is easy to implement and already delivers the strongest average
self-supervised rank in Table~\ref{tab:main}. It also clarifies which pieces
are load-bearing. Collapse prevention and target isolation matter more than
the exact structural-descriptor pathway, which is useful guidance for
follow-up implementations.

\section{Efficiency Analysis}
\label{sec:efficiency}

Accuracy alone does not reveal how the two masking units behave as graphs
grow. This section shows that PatchJEPA turns the same latent-prediction
objective into a much cheaper training procedure on large graphs, which is a
practical advantage of the dual-variant design.

\paragraph{Wall-clock cost across graph sizes.} On Coauthor-Physics (34K
nodes) all methods finish in under three minutes. On ogbn-arxiv (169K nodes)
PatchJEPA and every baseline finish in $24$--$41$ minutes, while NodeJEPA
takes $\sim$8.8 hours ($31{,}582$s mean over 5 seeds) under the uncached
$k$-hop expansion used in our reference implementation. The same objective
therefore supports two operating points: fine-grained node-level masking when
neighborhood structure is the priority, and cached patch masking when
wall-clock time on a large graph is the priority. This flexibility is useful
in practice. A practitioner can start with NodeJEPA on medium graphs, where
it is often the strongest or near-strongest self-supervised method, and switch
to PatchJEPA when scaling to hub-heavy networks without redesigning the loss
or encoder. The appendix plots accuracy against wall-clock time and gives
the full timing table.

\begin{figure}[t]
\centering
\includegraphics[width=0.78\columnwidth]{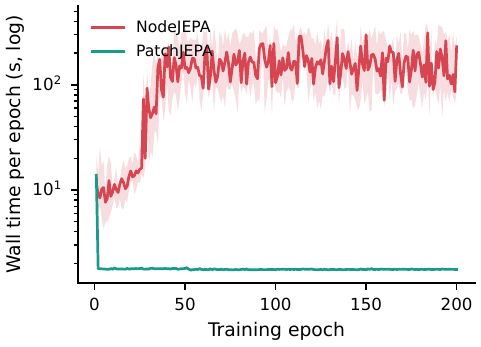}
\caption{Per-epoch wall-clock time on ogbn-arxiv (median and min-max range
over 5 seeds for NodeJEPA, 3 for PatchJEPA). NodeJEPA's cost jumps once the
masking curriculum reaches 2-hop neighborhoods around epoch 25-50 and stays
volatile; PatchJEPA is flat because its partition is cached.}
\label{fig:epoch}
\end{figure}

\paragraph{Why PatchJEPA stays flat.} NodeJEPA refreshes $k$-hop masks every
step (Section~\ref{sec:method}). On hub-heavy citation graphs such as
ogbn-arxiv, expanding to $2$ hops around a high-degree seed can touch a large
fraction of the graph. Figure~\ref{fig:epoch} shows the resulting jump in
median epoch time from about $11$s in the first $25$ epochs to about $147$s
thereafter (p95 $320.7$s). PatchJEPA stays near $1.8$s per epoch across all
$200$ epochs, a $75$--$180\times$ advantage, because its METIS partition is
computed once and cached (Section~\ref{sec:patchjepa}). Growing the masking
unit then no longer requires searching the graph again. This is the main
efficiency benefit of the patch-level instantiation. In other words, the
expensive search is paid once as preprocessing, after which every training
step only samples cached patches. That amortization is what keeps PatchJEPA
competitive with standard baselines on wall-clock time while still optimizing
a latent-prediction loss.

\paragraph{Practical takeaway.} Caching or degree-capping neighborhood search
\citep{hamilton2017graphsage} can further accelerate the node-level variant
without changing its masking granularity. Figure~\ref{fig:epoch} reports the
reference design as specified. Under a hard wall-clock budget, PatchJEPA is
already a strong drop-in alternative that preserves the JEPA objective while
keeping training time competitive with standard baselines
(Table~\ref{tab:main}). The dual design therefore gives practitioners a clear
accuracy--efficiency choice rather than a single fixed operating point.

Equally important, the efficiency gap is not a weakness of latent prediction
itself. It is a property of the masking unit. Once the partition is cached,
PatchJEPA matches baseline wall-clock cost while remaining among the top
self-supervised methods on accuracy. This is the main systems-level advantage
of studying the two variants together. NodeJEPA shows what fine-grained
neighborhood prediction can achieve. PatchJEPA shows how to keep that
objective practical on large graphs. Relative to prior graph SSL methods that
expose only one masking or augmentation recipe, the side-by-side comparison
makes the accuracy--efficiency frontier explicit and actionable. We therefore
recommend reporting both accuracy and wall-clock cost when proposing new
masking schemes for graph JEPA, so that scalability claims can be checked
directly rather than inferred from asymptotic arguments alone.

\section{Limitations}
\label{sec:limitations}

Our main suite covers homophilous citation, co-purchase, and co-authorship
graphs. Extending the same protocol to weaker-homophily settings
\citep{zhu2020heterophily,pei2020geomgcn} and other encoders
\citep{velickovic2018gat,xu2019gin,gilmer2017mpnn} is a natural next step.
Adaptive regularization across datasets and tighter caching of $k$-hop
neighborhoods are also promising extensions suggested by
Sections~\ref{sec:ablation} and~\ref{sec:efficiency}. Neither extension
changes the core claim that latent neighborhood prediction is a strong
self-supervised objective under a matched protocol. The appendix provides
additional diagnostics.

\section{Conclusion}

\looseness=1
We presented NodeJEPA and PatchJEPA for node-level graph self-supervised
learning. Both predict latent targets rather than reconstructing inputs, and
together they show that JEPA-style objectives are competitive with strong
contrastive and generative baselines under a matched protocol. Across five
benchmarks the two variants attain the best average self-supervised rank,
transfer well in few-shot settings, and offer a practical choice between
fine-grained $k$-hop masking and efficient cached patch masking. Ablations
further support context isolation, cosine prediction, and
variance-covariance regularization as useful default ingredients of the
recipe. We view this dual design as a concrete way to bring latent prediction
to graphs at both medium and large scale without giving up a matched fair
comparison against strong baselines. Code, configurations, and evaluation
scripts are publicly available at
\url{https://github.com/OliverZ-dot/Node-Jepa} to support exact reproduction
of the reported tables and figures. We hope this package makes follow-up work
on graph JEPA easier to build and compare under the same evaluation protocol
used in this paper. That shared protocol is what makes the reported gains
attributable to the pretraining objective rather than to unequal encoder
choices or unequal split settings.

\appendix

\section{Appendix Overview}
\label{app:overview}

This appendix expands the main paper with algorithmic pseudocode, full
protocol and hyperparameter details, complete statistical and few-shot
tables, representation-geometry diagnostics, and efficiency breakdowns. All
numbers come from the same experimental suite reported in the main paper.
NodeJEPA and PatchJEPA jointly achieve the best average self-supervised rank
(2.4) across five node-classification benchmarks.

\section{Algorithmic Description}
\label{app:algo}

\noindent\textbf{Algorithm 1} NodeJEPA, one training step.
\label{alg:nodejepa}\vspace{0.3em}
\begin{algorithmic}[1]
\Require graph $G=(V,E)$, features $X$, context encoder $f_\theta$, target
encoder $f_\xi$, predictor $g_\phi$, momentum $m$, mask ratio $r_t$, hop
radius $k_t$
\State Sample seeds and expand to $k_t$-hop neighborhoods until
$|\mathcal{T}|\approx r_t|V|$
\State Replace features of nodes in $\mathcal{T}$ with a learned mask token
\State $H^{\text{ctx}}\gets f_\theta(\tilde{X},E)$;
$H^{\text{tgt}}\gets\textsc{StopGrad}(f_\xi(X,E))$
\State Condition predictor on structural descriptors
$d_i$ (PageRank, degree, clustering, spectral coordinates)
\State $\mathcal{L}_{\text{pred}}\gets\frac{1}{|\mathcal{T}|}\sum_{i\in\mathcal{T}}(1-\cos(H^{\text{pred}}_i,H^{\text{tgt}}_i))$
\State Add VICReg-style variance/covariance and sketched isotropic-Gaussian
penalties; update $\theta$; EMA-update $\xi$
\end{algorithmic}

\vspace{0.6em}
\noindent\textbf{Algorithm 2} PatchJEPA, one training step.
\label{alg:patchjepa}\vspace{0.3em}
\begin{algorithmic}[1]
\Require METIS patches $\{P_k\}$ (cached), encoders $f_\theta,f_\xi$,
predictor $g_\phi$, momentum $m$
\State Sample 1 context patch and 4 target patches; expand each by one hop
\State Mean-pool context/target patch embeddings from $f_\theta$ / stop-grad
$f_\xi$
\State Predict target patch embeddings conditioned on patch descriptors;
add hyperbola-matching and SIGReg terms \citep{skenderi2023graphjepa}
\State Update $\theta$; EMA-update $\xi$
\end{algorithmic}

\section{Dataset Statistics}
\label{app:datasets}

Table~\ref{tab:app-datasets} lists node, edge, feature, and class counts for all five benchmarks.

\begin{table*}[t]
\centering
\small
\setlength{\tabcolsep}{4pt}
\begin{tabular}{lrrrrl}
\toprule
Dataset & Nodes & Edges & Feat. & Classes & Train/Val/Test \\
\midrule
Amazon-Computers & 13{,}752 & 491{,}722 & 767 & 10 & 200/300/rest \\
Amazon-Photo & 7{,}650 & 238{,}162 & 745 & 8 & 160/240/rest \\
Coauthor-CS & 18{,}333 & 163{,}788 & 6{,}805 & 15 & 300/439/rest \\
Coauthor-Physics & 34{,}493 & 495{,}924 & 8{,}415 & 5 & 100/150/rest \\
ogbn-arxiv & 169{,}343 & 2{,}315{,}598 & 128 & 40 & OGB official \\
\bottomrule
\end{tabular}
\caption{Dataset statistics. Non-OGB splits use $20$/$30$ labels per class for train/validation \citep{shchur2018pitfalls}; ogbn-arxiv uses the official split \citep{hu2020ogb}.}
\label{tab:app-datasets}
\end{table*}

\section{Hyperparameters}
\label{app:hparams}

Tables~\ref{tab:app-shared} and~\ref{tab:app-method} give the shared and method-specific settings used for every run.

\begin{table}[t]
\centering
\small
\begin{tabular}{ll}
\toprule
Hyperparameter & Value \\
\midrule
Encoder & GCN, 3 layers, 256 hidden \\
Optimizer & AdamW, lr $10^{-3}$, weight decay $10^{-4}$ \\
LR schedule & Cosine to $10^{-6}$; grad clip $1.0$ \\
Epochs & 300 (Amazon/Coauthor), 200 (arxiv) \\
Seeds & $\{42,43,44,45,46\}$ \\
\bottomrule
\end{tabular}
\caption{Shared training hyperparameters across all seven methods.}
\label{tab:app-shared}
\end{table}

\begin{table*}[t]
\centering
\small
\begin{tabular}{lp{11cm}}
\toprule
Method & Method-specific hyperparameters \\
\midrule
NodeJEPA & restricted GCN predictor (3$\times$256); EMA $0.996\to0.999$;
mask $0.2\to0.5$ over 50 epochs; hops $1\to2$;
$\lambda_{\mathrm{var}}=\lambda_{\mathrm{cov}}=0.2$,
$\lambda_{\mathrm{sig}}=0.02$ (256 slices) \\
PatchJEPA & $n_{\mathrm{patches}}=64/64/64/128/256$; 1 context / 4 targets;
1-hop expansion; cached METIS; EMA $0.996$; SIGReg $0.02$ \\
DGI & bilinear discriminator; feature-shuffle corruption \\
GraphMAE & mask $0.5$; scaled cosine error $\alpha=2.0$ \\
BGRL & edge/feature drop $0.2$; EMA $0.996$ \\
CCA-SSG & edge/feature drop $0.2$; $\lambda=10^{-3}$; subsample $8192$ \\
Sup.\ GCN & dropout $0.5$; cross-entropy \\
\bottomrule
\end{tabular}
\caption{Method-specific hyperparameters, fixed across all five datasets.}
\label{tab:app-method}
\end{table*}

\section{Hardware and Evaluation Protocol}
\label{app:hardware}
\label{app:protocol}

All runs use one NVIDIA A100 80GB GPU, Python 3.8.10, PyTorch 2.4.1
(CUDA 12.1), and PyTorch Geometric $\ge$2.4. After pretraining we freeze the
encoder and fit multinomial logistic regression (scikit-learn). Few-shot
probes use $k\in\{5,10,20\}$ labels per class. Significance uses paired
$t$-tests on shared seeds (Welch when seed sets differ), threshold $0.05$.
Average rank is computed among the six self-supervised methods only;
NodeJEPA and PatchJEPA both obtain $2.4$.

\section{Main Results, Visualized}
\label{app:mainfig}

\begin{figure*}[t]
\centering
\includegraphics[width=\textwidth]{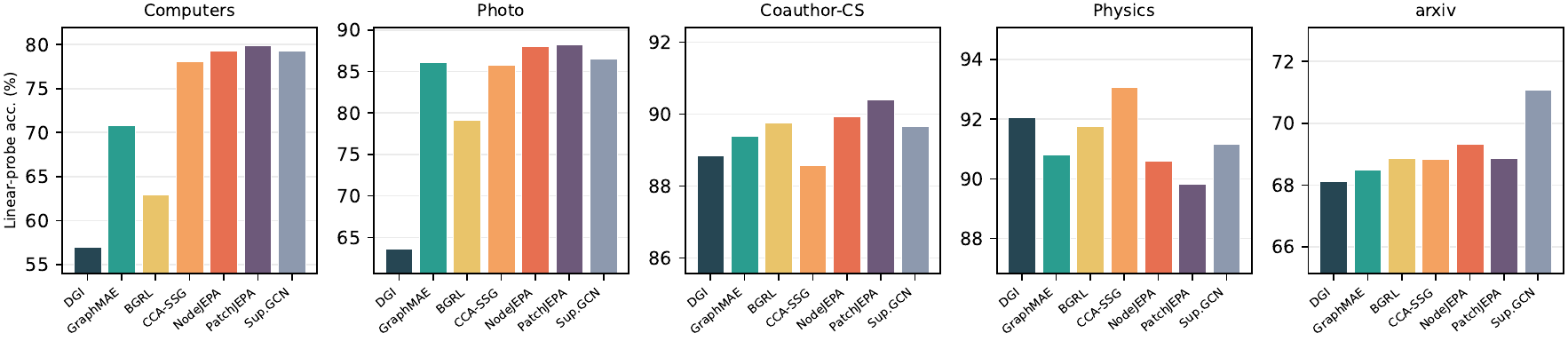}
\caption{Linear-probe accuracy across all five benchmarks (same data as Table~\ref{tab:main} of the main paper).}
\label{fig:app-main}
\end{figure*}

Figure~\ref{fig:app-main} and Table~\ref{tab:app-main} reproduce the main linear-probe numbers as a bar chart and a compact table.

\begin{table*}[t]
\centering
\scriptsize
\setlength{\tabcolsep}{3pt}
\begin{tabular}{lcccccc}
\toprule
Method & Computers & Photo & Coauthor-CS & Physics & arxiv & Avg.\ Rank \\
\midrule
DGI & $56.96{\pm}18.09$ & $63.62{\pm}13.00$ & $88.85{\pm}0.35$ & $92.04{\pm}0.23$ & $68.13{\pm}0.02$ & 5.0 \\
GraphMAE & $70.76{\pm}2.92$ & $86.05{\pm}0.52$ & $89.39{\pm}0.06$ & $90.80{\pm}0.39$ & $68.49{\pm}0.21$ & 4.0 \\
BGRL & $62.95{\pm}3.01$ & $79.18{\pm}3.15$ & $89.76{\pm}0.15$ & $91.76{\pm}0.77$ & $68.87{\pm}0.20$ & 3.6 \\
CCA-SSG & $78.10{\pm}0.35$ & $85.75{\pm}0.53$ & $88.56{\pm}0.07$ & $93.06{\pm}0.15$ & $68.85{\pm}0.19$ & 3.6 \\
Sup.\ GCN & $79.26{\pm}0.31$ & $86.48{\pm}0.26$ & $89.67{\pm}0.17$ & $91.17{\pm}0.11$ & $71.09{\pm}0.05$ & -- \\
\midrule
NodeJEPA & $79.23{\pm}1.27$ & $87.98{\pm}0.58$ & $89.93{\pm}0.37$ & $90.60{\pm}0.62$ & $69.31{\pm}0.16$ & \textbf{2.4} \\
PatchJEPA & $79.90{\pm}0.86$ & $88.27{\pm}0.46$ & $90.40{\pm}0.24$ & $89.82{\pm}1.33$ & $68.86{\pm}0.20$ & \textbf{2.4} \\
\bottomrule
\end{tabular}
\caption{Linear-probe accuracy (\%, mean~$\pm$~std over 5 seeds), reproduced from Table~\ref{tab:main} of the main paper.}
\label{tab:app-main}
\end{table*}

\section{Pairwise Wins and Deltas}
\label{app:wins}

\begin{figure*}[t]
\centering
\begin{minipage}[t]{0.32\textwidth}
\centering
\includegraphics[width=\linewidth]{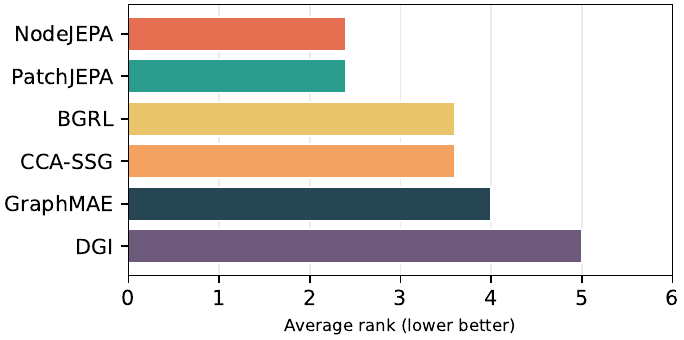}
\end{minipage}\hfill
\begin{minipage}[t]{0.32\textwidth}
\centering
\includegraphics[width=\linewidth]{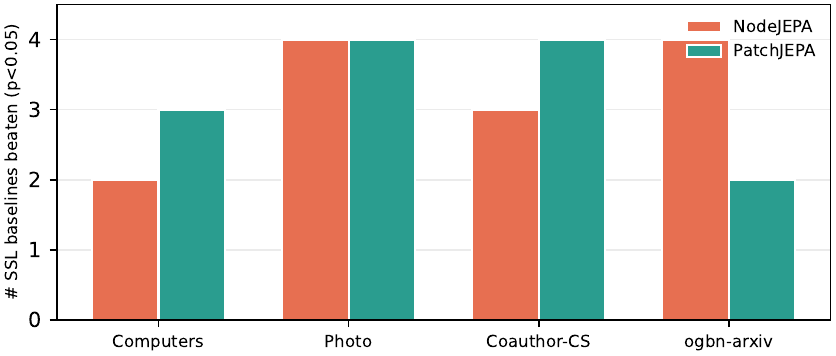}
\end{minipage}\hfill
\begin{minipage}[t]{0.32\textwidth}
\centering
\includegraphics[width=\linewidth]{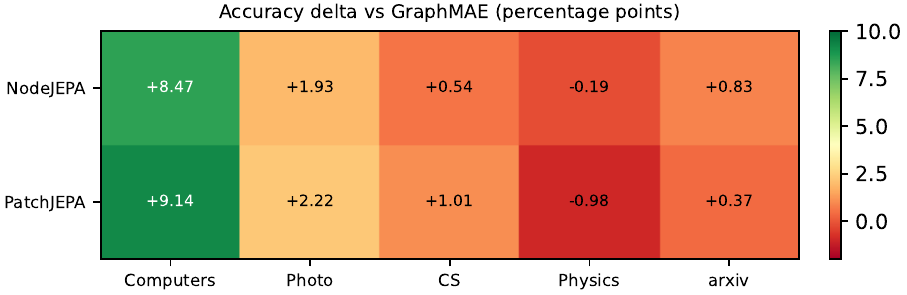}
\end{minipage}
\caption{Left: average rank (lower better). Middle: SSL baselines beaten at $p<0.05$. Right: accuracy delta vs.\ GraphMAE (pp).}
\label{fig:app-rankwins}
\label{fig:app-rank}
\label{fig:app-wins}
\label{fig:app-delta}
\end{figure*}

Figure~\ref{fig:app-rankwins} summarizes average rank, significant wins against SSL baselines, and accuracy deltas versus GraphMAE.

\section{Extended Significance Testing}
\label{app:significance}

Tables~\ref{tab:app-sig-node} and~\ref{tab:app-sig-patch} report the full
pairwise $t$-tests underlying Section~\ref{sec:experiments} of the main paper.

\begin{table*}[t]
\centering
\scriptsize
\setlength{\tabcolsep}{3pt}
\begin{tabular}{llrrrrl}
\toprule
Dataset & Baseline & Mean NJ & Mean base & Diff & $p$ & Test \\
\midrule
Computers & DGI & 79.23 & 56.96 & +22.27 & 0.071 & paired \\
Computers & GraphMAE & 79.23 & 70.76 & +8.47 & 0.007** & paired \\
Computers & BGRL & 79.23 & 62.95 & +16.28 & 0.001** & paired \\
Computers & CCA-SSG & 79.23 & 78.10 & +1.13 & 0.143 & paired \\
Computers & Sup.\ GCN & 79.23 & 79.26 & -0.03 & 0.956 & paired \\
Photo & DGI & 87.98 & 63.62 & +24.35 & 0.019* & paired \\
Photo & GraphMAE & 87.98 & 86.05 & +1.93 & 0.003** & paired \\
Photo & BGRL & 87.98 & 79.18 & +8.79 & 0.004** & paired \\
Photo & CCA-SSG & 87.98 & 85.75 & +2.23 & 0.010* & paired \\
Photo & Sup.\ GCN & 87.98 & 86.48 & +1.50 & 0.016* & paired \\
Coauthor-CS & DGI & 89.93 & 88.85 & +1.08 & 0.012* & paired \\
Coauthor-CS & GraphMAE & 89.93 & 89.39 & +0.54 & 0.038* & paired \\
Coauthor-CS & BGRL & 89.93 & 89.76 & +0.17 & 0.490 & paired \\
Coauthor-CS & CCA-SSG & 89.93 & 88.56 & +1.37 & 0.002** & paired \\
Coauthor-CS & Sup.\ GCN & 89.93 & 89.67 & +0.26 & 0.319 & paired \\
Physics & DGI & 90.60 & 92.04 & -1.43 & 0.014* & paired \\
Physics & GraphMAE & 90.60 & 90.80 & -0.19 & 0.439 & paired \\
Physics & BGRL & 90.60 & 91.76 & -1.15 & 0.156 & paired \\
Physics & CCA-SSG & 90.60 & 93.06 & -2.46 & 0.002** & paired \\
Physics & Sup.\ GCN & 90.60 & 91.17 & -0.57 & 0.189 & paired \\
arxiv & DGI & 69.31 & 68.13 & +1.18 & 0.0001** & Welch \\
arxiv & GraphMAE & 69.31 & 68.49 & +0.83 & 0.001** & paired \\
arxiv & BGRL & 69.31 & 68.87 & +0.45 & 0.021* & paired \\
arxiv & CCA-SSG & 69.31 & 68.85 & +0.46 & 0.044* & paired \\
arxiv & Sup.\ GCN & 69.31 & 71.09 & -1.78 & 0.0001** & paired \\
\bottomrule
\end{tabular}
\caption{NodeJEPA vs.\ every baseline. `*' $p<0.05$, `**' $p<0.01$.}
\label{tab:app-sig-node}
\end{table*}

\begin{table*}[t]
\centering
\scriptsize
\setlength{\tabcolsep}{3pt}
\begin{tabular}{llrrrrl}
\toprule
Dataset & Baseline & Mean PJ & Mean base & Diff & $p$ & Test \\
\midrule
Computers & DGI & 79.90 & 56.96 & +22.94 & 0.065 & paired \\
Computers & GraphMAE & 79.90 & 70.76 & +9.14 & 0.003** & paired \\
Computers & BGRL & 79.90 & 62.95 & +16.95 & 0.0003** & paired \\
Computers & CCA-SSG & 79.90 & 78.10 & +1.80 & 0.004** & paired \\
Computers & Sup.\ GCN & 79.90 & 79.26 & +0.64 & 0.199 & paired \\
Photo & DGI & 88.27 & 63.62 & +24.64 & 0.020* & paired \\
Photo & GraphMAE & 88.27 & 86.05 & +2.22 & 0.0001** & paired \\
Photo & BGRL & 88.27 & 79.18 & +9.08 & 0.005** & paired \\
Photo & CCA-SSG & 88.27 & 85.75 & +2.52 & 0.002** & paired \\
Photo & Sup.\ GCN & 88.27 & 86.48 & +1.79 & 0.005** & paired \\
Coauthor-CS & DGI & 90.40 & 88.85 & +1.56 & 0.002** & paired \\
Coauthor-CS & GraphMAE & 90.40 & 89.39 & +1.01 & 0.001** & paired \\
Coauthor-CS & BGRL & 90.40 & 89.76 & +0.64 & 0.022* & paired \\
Coauthor-CS & CCA-SSG & 90.40 & 88.56 & +1.84 & 0.0001** & paired \\
Coauthor-CS & Sup.\ GCN & 90.40 & 89.67 & +0.73 & 0.015* & paired \\
Physics & DGI & 89.82 & 92.04 & -2.22 & 0.030* & paired \\
Physics & GraphMAE & 89.82 & 90.80 & -0.98 & 0.227 & paired \\
Physics & BGRL & 89.82 & 91.76 & -1.93 & 0.008** & paired \\
Physics & CCA-SSG & 89.82 & 93.06 & -3.24 & 0.010** & paired \\
Physics & Sup.\ GCN & 89.82 & 91.17 & -1.35 & 0.115 & paired \\
arxiv & DGI & 68.86 & 68.13 & +0.72 & 0.002** & Welch \\
arxiv & GraphMAE & 68.86 & 68.49 & +0.37 & 0.002** & paired \\
arxiv & BGRL & 68.86 & 68.87 & -0.01 & 0.922 & paired \\
arxiv & CCA-SSG & 68.86 & 68.85 & +0.01 & 0.964 & paired \\
arxiv & Sup.\ GCN & 68.86 & 71.09 & -2.23 & 0.0001** & paired \\
\bottomrule
\end{tabular}
\caption{PatchJEPA vs.\ every baseline.}
\label{tab:app-sig-patch}
\end{table*}

\section{Extended Few-Shot Results}
\label{app:fewshot}

\begin{figure*}[t]
\centering
\includegraphics[width=\textwidth]{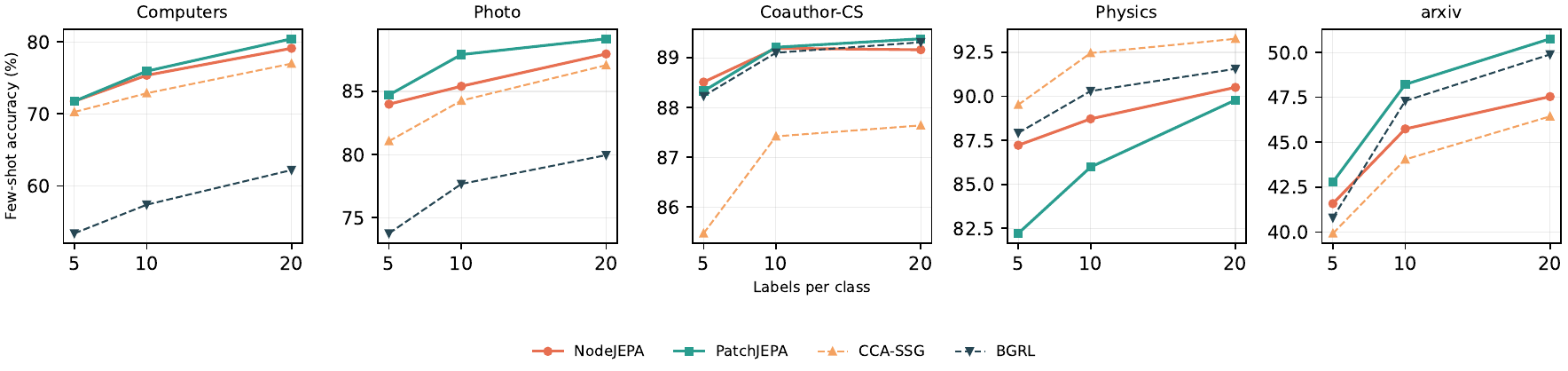}
\caption{Few-shot accuracy vs.\ labels per class on all five datasets.}
\label{fig:app-fewshot}
\end{figure*}

Figure~\ref{fig:app-fewshot} and Table~\ref{tab:app-fewshot} give few-shot curves and numbers for all five datasets and three label budgets.

\begin{table*}[t]
\centering
\scriptsize
\setlength{\tabcolsep}{3pt}
\begin{tabular}{llrrr}
\toprule
Method & Dataset & 5/class & 10/class & 20/class \\
\midrule
NodeJEPA & Computers & 71.75$\pm$2.88 & 75.37$\pm$1.48 & 79.12$\pm$0.45 \\
NodeJEPA & Photo & 83.99$\pm$3.02 & 85.41$\pm$2.52 & 87.97$\pm$0.84 \\
NodeJEPA & Coauthor-CS & 88.51$\pm$0.73 & 89.19$\pm$1.17 & 89.16$\pm$0.99 \\
NodeJEPA & Physics & 87.22$\pm$1.63 & 88.72$\pm$1.76 & 90.51$\pm$1.82 \\
NodeJEPA & arxiv & 41.58$\pm$1.16 & 45.74$\pm$1.94 & 47.54$\pm$0.52 \\
PatchJEPA & Computers & 71.76$\pm$4.14 & 75.94$\pm$1.88 & 80.44$\pm$0.77 \\
PatchJEPA & Photo & 84.72$\pm$0.47 & 87.92$\pm$2.01 & 89.17$\pm$0.86 \\
PatchJEPA & Coauthor-CS & 88.33$\pm$0.39 & 89.21$\pm$0.44 & 89.38$\pm$0.60 \\
PatchJEPA & Physics & 82.21$\pm$2.18 & 85.98$\pm$4.24 & 89.79$\pm$2.46 \\
PatchJEPA & arxiv & 42.78$\pm$1.70 & 48.22$\pm$0.64 & 50.76$\pm$0.73 \\
DGI & Computers & 51.14$\pm$14.00 & 55.39$\pm$16.22 & 63.41$\pm$13.35 \\
DGI & Photo & 44.38$\pm$12.76 & 55.84$\pm$12.36 & 64.89$\pm$9.19 \\
DGI & Coauthor-CS & 84.88$\pm$1.38 & 86.67$\pm$0.51 & 87.84$\pm$0.99 \\
DGI & Physics & 87.64$\pm$2.15 & 91.19$\pm$1.14 & 92.42$\pm$0.91 \\
GraphMAE & Computers & 58.93$\pm$5.32 & 62.28$\pm$5.42 & 68.85$\pm$4.95 \\
GraphMAE & Photo & 81.64$\pm$2.25 & 83.02$\pm$1.43 & 85.28$\pm$1.58 \\
GraphMAE & Coauthor-CS & 89.68$\pm$0.26 & 90.23$\pm$0.64 & 90.54$\pm$0.33 \\
GraphMAE & Physics & 86.84$\pm$2.04 & 87.90$\pm$2.98 & 89.93$\pm$2.28 \\
GraphMAE & arxiv & 39.17$\pm$1.66 & 46.89$\pm$1.00 & 49.98$\pm$0.94 \\
BGRL & Computers & 53.38$\pm$3.39 & 57.37$\pm$3.10 & 62.19$\pm$3.23 \\
BGRL & Photo & 73.75$\pm$6.26 & 77.67$\pm$2.11 & 79.95$\pm$1.32 \\
BGRL & Coauthor-CS & 88.23$\pm$0.81 & 89.10$\pm$1.06 & 89.31$\pm$0.99 \\
BGRL & Physics & 87.92$\pm$0.82 & 90.30$\pm$0.92 & 91.56$\pm$1.23 \\
BGRL & arxiv & 40.79$\pm$1.52 & 47.30$\pm$1.25 & 49.89$\pm$0.51 \\
CCA-SSG & Computers & 70.26$\pm$3.97 & 72.87$\pm$2.48 & 76.99$\pm$1.40 \\
CCA-SSG & Photo & 81.06$\pm$3.28 & 84.27$\pm$2.47 & 87.08$\pm$1.02 \\
CCA-SSG & Coauthor-CS & 85.47$\pm$0.82 & 87.42$\pm$0.60 & 87.64$\pm$0.70 \\
CCA-SSG & Physics & 89.52$\pm$1.53 & 92.46$\pm$0.58 & 93.27$\pm$0.52 \\
CCA-SSG & arxiv & 39.92$\pm$2.50 & 44.04$\pm$1.61 & 46.44$\pm$0.39 \\
Sup.\ GCN & Computers & 80.51$\pm$3.38 & 80.91$\pm$2.35 & 82.90$\pm$1.39 \\
Sup.\ GCN & Photo & 88.11$\pm$2.30 & 89.56$\pm$1.51 & 89.71$\pm$1.27 \\
Sup.\ GCN & Coauthor-CS & 89.19$\pm$0.44 & 89.27$\pm$0.70 & 89.42$\pm$0.65 \\
Sup.\ GCN & Physics & 90.17$\pm$1.52 & 91.45$\pm$0.85 & 92.27$\pm$0.68 \\
Sup.\ GCN & arxiv & 52.90$\pm$1.42 & 53.33$\pm$1.00 & 53.17$\pm$0.65 \\
\bottomrule
\end{tabular}
\caption{Few-shot linear-probe accuracy (\%).}
\label{tab:app-fewshot}
\end{table*}

\section{Ablations}
\label{app:ablation}

\begin{figure*}[t]
\centering
\includegraphics[width=0.85\textwidth]{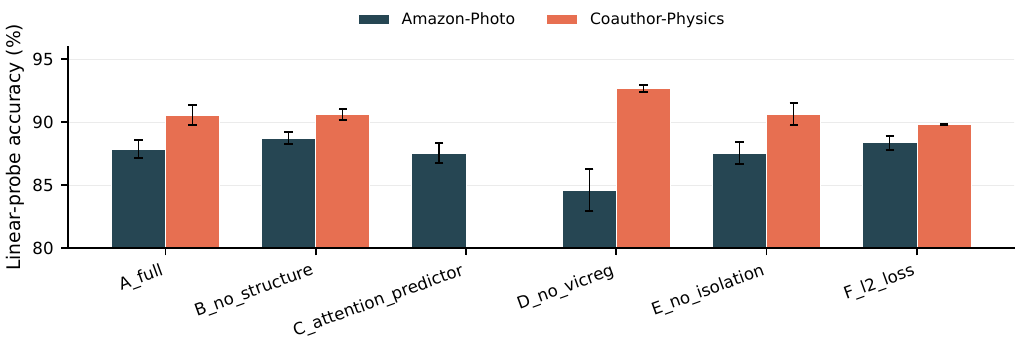}\vspace{0.6em}
\small
\begin{tabular}{lcc}
\toprule
Variant & Amazon-Photo & Coauthor-Phy. \\
\midrule
A: full & 87.87 $\pm$ 0.69 & 90.57 $\pm$ 0.80 \\
B: no structure & 88.74 $\pm$ 0.49 & 90.61 $\pm$ 0.43 \\
C: attn.\ predictor & 87.53 $\pm$ 0.78 & 90.18 $\pm$ 0.64 \\
D: no VICReg & 84.61 $\pm$ 1.67 & 92.67 $\pm$ 0.29 \\
E: no isolation & 87.54 $\pm$ 0.85 & 90.65 $\pm$ 0.88 \\
F: $\ell_2$ loss & 88.36 $\pm$ 0.56 & 89.79 $\pm$ 0.05 \\
\bottomrule
\end{tabular}
\caption{NodeJEPA ablation results on Amazon-Photo and Coauthor-Physics.}
\label{fig:app-ablation}
\captionof{table}{NodeJEPA ablations (3 seeds), matching the main paper.}
\label{tab:app-ablation}
\end{figure*}

Table~\ref{tab:app-ablation} and Figure~\ref{fig:app-ablation} match the
main-paper ablation suite (3 seeds), including the attention predictor at
$90.18\pm0.64$ on Coauthor-Physics.

Default recipe: isolate targets, cosine prediction, restricted
message-passing predictor, and VICReg-style regularization. Structural
descriptors remain an optional conditioning channel. Cross-attention (C)
matches the default closely ($90.18$ vs.\ $90.57$ on Physics).

\section{Representation-Collapse Diagnostics}
\label{app:collapse}

\begin{figure*}[t]
\centering
\includegraphics[width=\textwidth]{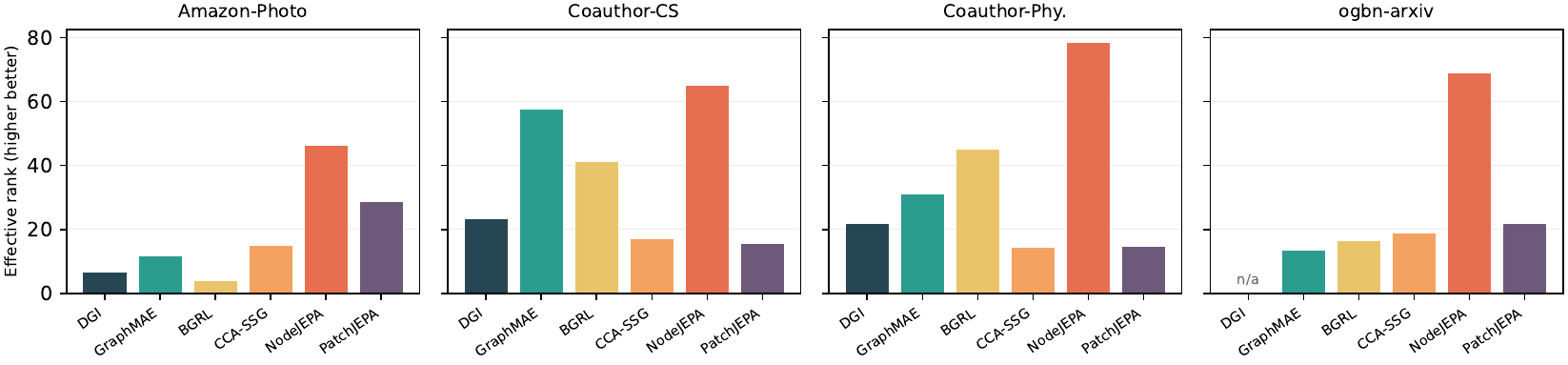}\vspace{0.3em}
\includegraphics[width=\textwidth]{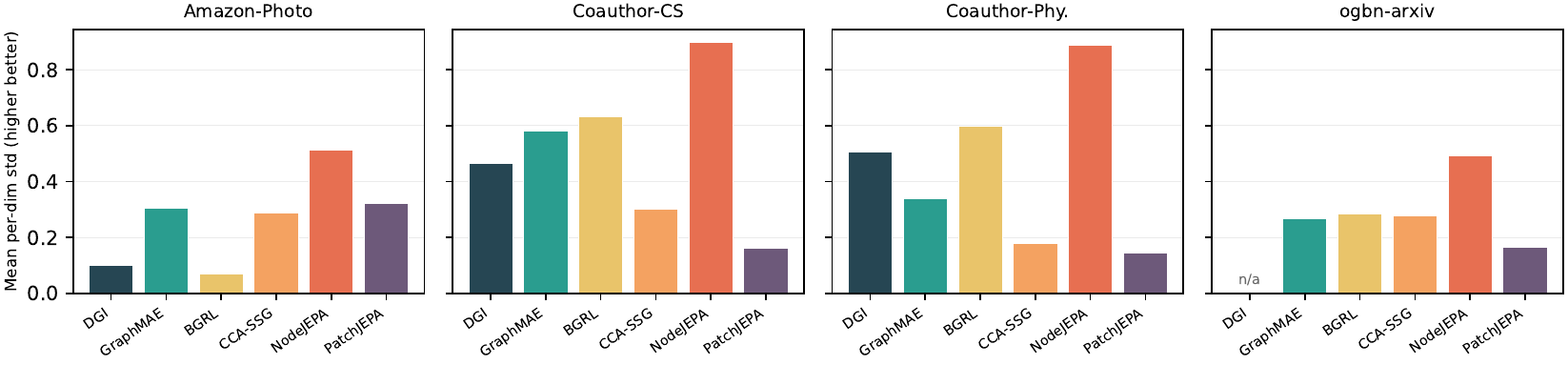}
\caption{Top: effective rank of frozen embeddings. Bottom: mean per-dimension standard deviation. NodeJEPA leads self-supervised methods on every logged dataset.}
\label{fig:app-collapse}
\label{fig:app-meanstd}
\end{figure*}

Figure~\ref{fig:app-collapse} and Tables~\ref{tab:app-collapse}--\ref{tab:app-collapse-full} report effective rank, mean per-dimension standard deviation, and participation ratio across datasets.

\begin{table*}[t]
\centering
\small
\begin{tabular}{lrrrr}
\toprule
Method & Photo & Coauthor-CS & Coauthor-Phy. & arxiv \\
\midrule
DGI & 6.7 & 23.5 & 21.9 & -- \\
GraphMAE & 11.8 & 57.7 & 31.1 & 13.4 \\
BGRL & 3.9 & 41.1 & 45.1 & 16.4 \\
CCA-SSG & 14.9 & 17.0 & 14.4 & 18.9 \\
Supervised GCN & 47.4 & 58.0 & 28.6 & 99.6 \\
\midrule
NodeJEPA & \textbf{46.3} & \textbf{65.2} & \textbf{78.5} & \textbf{68.9} \\
PatchJEPA & 28.6 & 15.7 & 14.6 & 22.0 \\
\bottomrule
\end{tabular}
\caption{Effective rank (higher means less collapse).}
\label{tab:app-collapse}
\vspace{0.8em}
\scriptsize
\begin{tabular}{llrrr}
\toprule
Dataset & Method & EffRank & MeanStd & PartRatio \\
\midrule
Photo & NodeJEPA & $46.3{\pm}5.6$ & $0.514{\pm}0.112$ & $1.87{\pm}0.46$ \\
Photo & PatchJEPA & $28.6{\pm}5.1$ & $0.323{\pm}0.040$ & $1.27{\pm}0.08$ \\
Photo & GraphMAE & $11.8{\pm}2.7$ & $0.304{\pm}0.071$ & $1.27{\pm}0.13$ \\
Photo & CCA-SSG & $14.9{\pm}1.0$ & $0.289{\pm}0.015$ & $1.25{\pm}0.03$ \\
Photo & BGRL & $3.9{\pm}1.4$ & $0.070{\pm}0.030$ & $1.01{\pm}0.01$ \\
CS & NodeJEPA & $65.2{\pm}2.6$ & $0.898{\pm}0.005$ & $7.84{\pm}0.33$ \\
CS & GraphMAE & $57.7{\pm}0.7$ & $0.580{\pm}0.005$ & $2.27{\pm}0.04$ \\
CS & BGRL & $41.1{\pm}1.7$ & $0.633{\pm}0.006$ & $2.67{\pm}0.06$ \\
CS & PatchJEPA & $15.7{\pm}1.9$ & $0.161{\pm}0.023$ & $1.06{\pm}0.02$ \\
Physics & NodeJEPA & $78.5{\pm}3.9$ & $0.887{\pm}0.008$ & $7.18{\pm}0.53$ \\
Physics & BGRL & $45.1{\pm}3.6$ & $0.598{\pm}0.018$ & $2.36{\pm}0.14$ \\
Physics & GraphMAE & $31.1{\pm}3.2$ & $0.339{\pm}0.039$ & $1.29{\pm}0.08$ \\
Physics & PatchJEPA & $14.6{\pm}0.1$ & $0.144{\pm}0.005$ & $1.04{\pm}0.00$ \\
arxiv & NodeJEPA & $68.9{\pm}8.9$ & $0.492{\pm}0.057$ & $1.75{\pm}0.27$ \\
arxiv & PatchJEPA & $22.0{\pm}2.2$ & $0.166{\pm}0.013$ & $1.06{\pm}0.01$ \\
arxiv & CCA-SSG & $18.9{\pm}1.1$ & $0.276{\pm}0.016$ & $1.20{\pm}0.02$ \\
arxiv & BGRL & $16.4{\pm}3.2$ & $0.284{\pm}0.030$ & $1.21{\pm}0.05$ \\
\bottomrule
\end{tabular}
\caption{Expanded collapse diagnostics.}
\label{tab:app-collapse-full}
\end{table*}

\clearpage
\section{Efficiency Analysis}
\label{app:efficiency}

\begin{figure*}[t]
\centering
\includegraphics[width=0.88\textwidth]{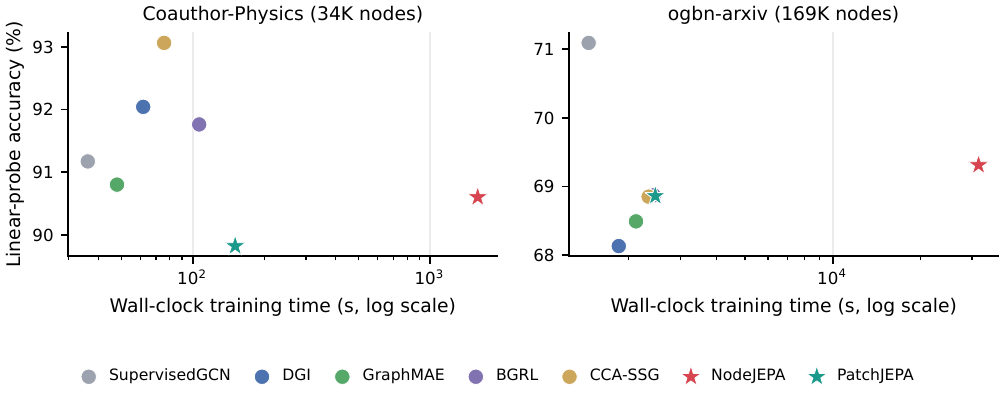}\vspace{0.2em}
\includegraphics[width=0.95\textwidth]{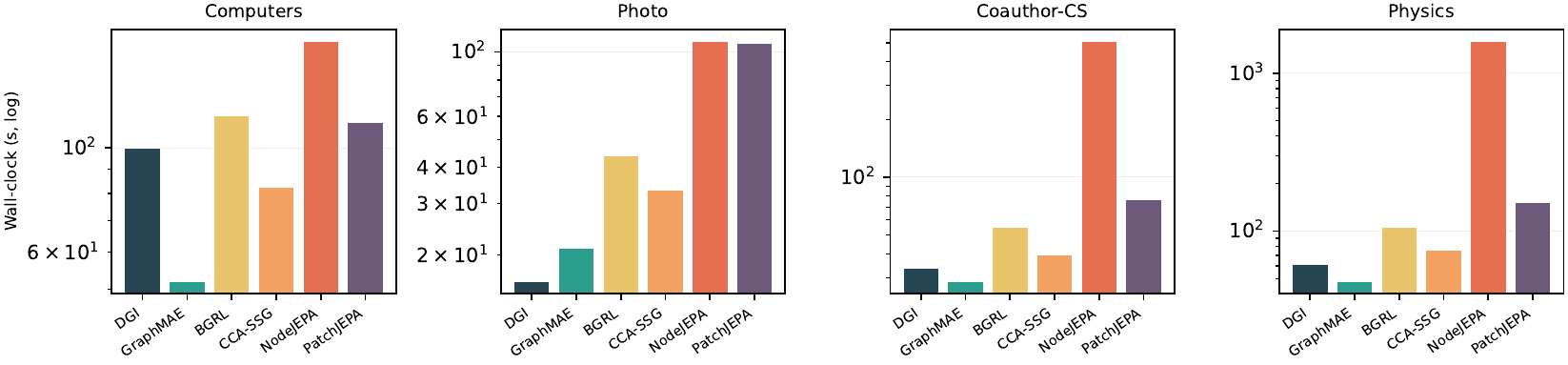}\vspace{0.2em}
\includegraphics[width=0.68\textwidth]{figures/fig_epoch_time.pdf}
\caption{Top: accuracy vs.\ wall-clock time. Middle: total training time on four medium graphs. Bottom: per-epoch time on ogbn-arxiv.}
\label{fig:app-tradeoff}
\label{fig:app-efficiency}
\label{fig:app-epoch}
\end{figure*}

Figure~\ref{fig:app-tradeoff} and Table~\ref{tab:app-efficiency} plot accuracy against wall-clock time and list mean training times on all five benchmarks, including per-epoch dynamics on ogbn-arxiv.

\begin{table*}[t]
\centering
\small
\begin{tabular}{lrrrrr}
\toprule
Method & Computers & Photo & Coauthor-CS & Physics & arxiv \\
\midrule
DGI & 99.8 & 16.2 & 33.7 & 61.8 & 1848.4 \\
GraphMAE & 52.0 & 21.0 & 28.7 & 47.9 & 2117.7 \\
BGRL & 117.0 & 44.0 & 54.6 & 106.3 & 2436.5 \\
CCA-SSG & 82.4 & 33.4 & 39.5 & 75.7 & 2339.2 \\
Sup.\ GCN & 41.1 & 15.2 & 13.4 & 36.1 & 1457.7 \\
NodeJEPA & 168.3 & 108.4 & 506.4 & 1585.5 & 31582.3 \\
PatchJEPA & 113.5 & 107.2 & 76.6 & 150.8 & 2466.3 \\
\bottomrule
\end{tabular}
\caption{Mean total training wall-clock time (seconds).}
\label{tab:app-efficiency}
\end{table*}

The dual-variant design offers two operating points: NodeJEPA for
fine-grained node-level geometry on medium graphs, and PatchJEPA for
cached-partition scaling on large hub-heavy graphs. On ogbn-arxiv, NodeJEPA's
median epoch time rises from about $11$s to about $147$s after the 2-hop
curriculum, while PatchJEPA stays near $1.8$s per epoch.

\clearpage
\section{Design Notes}
\label{app:design}

Structural descriptors (PageRank, degree, clustering, spectral coordinates)
are optional. The NodeJEPA curriculum raises mask ratio $0.2\to0.5$ and hop
radius $1\to2$ over the first 50 epochs. Context isolation replaces target
features with a mask token. Node-level and patch-level masking share the same
JEPA loss family and differ mainly in the masking unit.

\section{Additional Method Details}
\label{app:method-extra}

EMA targets use momentum $0.996\to0.999$ (NodeJEPA) or $0.996$ (PatchJEPA).
The predictive loss is mean cosine distance. SIGReg uses 256 random
projections with weight $0.02$. PatchJEPA mean-pools within patches only for
the pretraining target; probing still uses node-level encoder outputs.

\section{Broader Impact}
\label{app:impact}

This work studies self-supervised learning on standard public benchmarks and
does not introduce new sensitive data. Downstream deployments should audit
structural or attribute biases as appropriate. Identifying the $k$-hop-search
bottleneck and providing a cached patch alternative reduces compute on large
graphs.

\section{Reproducibility Notes}
\label{app:repro}

Seeds $42$--$46$, splits, and hyperparameters are fully specified above.
Code, configurations, scripts, and aggregated logs (MIT license) are
publicly available at \url{https://github.com/OliverZ-dot/Node-Jepa}.

\bibliography{references}

\end{document}